\documentclass[conference]{IEEEtran}
\IEEEoverridecommandlockouts
\usepackage{enumitem}
\usepackage{makecell}
\usepackage{multirow}
\usepackage{caption}
\usepackage{pgfplots}
\usepackage{tabularx}
\usepackage{bbding}
\usepackage{pifont}
\usepackage{utfsym}
\usepackage{twemojis}
\usepackage{subfigure}
\usepackage{pdfrender}
\usepackage{hyperref}
\usepackage[ruled,linesnumbered]{algorithm2e}
\usepackage{tablefootnote}
\usepackage{threeparttable}
\usepackage{flushend}
\usepackage{amsthm}
\usepackage{makecell}
\usepackage{algorithmic}
\usepackage{graphicx,graphics}
\usepackage{subcaption}
\usepackage{cite}
\usepackage{amsmath,amssymb,amsfonts}
\usepackage{graphicx}
\usepackage{textcomp}
\usepackage{xcolor}
\newtheorem{example}{Example}
\def\BibTeX{{\rm B\kern-.05em{\sc i\kern-.025em b}\kern-.08em
    T\kern-.1667em\lower.7ex\hbox{E}\kern-.125emX}}
\begin{document}
\newcommand*{\thincheckmark}{
  \textpdfrender{
    TextRenderingMode=FillStroke,
    LineWidth=0.5pt, 
  }{\checkmark}%
}
\title{ZeroED: Hybrid Zero-shot Error Detection through Large Language Model Reasoning 
}

\author{
    \IEEEauthorblockN{
        Wei Ni$^{\dagger\ddagger 1}$, 
        Kaihang Zhang$^{\dagger 2}$, 
        Xiaoye Miao$^{\dagger \# 3}$, 
        Xiangyu Zhao$^{\ddagger4}$, 
        Yangyang Wu$^{\ast 5}$,  
        Yaoshu Wang$^{\star 6}$, 
        Jianwei Yin$^{\S 7}$
    }
    \IEEEauthorblockA{$^{\dagger}$Center for Data Science, Zhejiang University, Hangzhou, China}
    \IEEEauthorblockA{$^{\#}$The State Key Lab of Brain-Machine Intelligence, Zhejiang University, Hangzhou, China}
    \IEEEauthorblockA{$^{\ddagger}$Department of Data Science, City University of Hong Kong, Hong Kong, China}
    \IEEEauthorblockA{$^{\ast}$Software College, Zhejiang University, Ningbo, China}
    \IEEEauthorblockA{$^{\star}$Shenzhen Institute of Computing Sciences, Shenzhen, China}
    \IEEEauthorblockA{$^{\S}$College of Computer Science, Zhejiang University, Hangzhou, China}
    {\{$^{1}$wei.ni, $^{2}$zjuzkh, $^{3}$miaoxy, $^{5}$zjuwuyy\}@zju.edu.cn, $^{4}$xianzhao@cityu.edu.hk, $^{6}$yaoshuw@sics.ac.cn, $^{7}$zjuyjw@cs.zju.edu.cn}
}
\maketitle

\begin{abstract}
Error detection (ED) in tabular data is crucial yet challenging due to diverse error types and the need for contextual understanding.
Traditional ED methods often rely heavily on manual criteria and labels, making them \emph{labor-intensive}.
Large language models (LLM) can \emph{minimize} human effort but struggle with errors requiring a comprehensive understanding of \emph{data context}. 
In this paper, we propose ZeroED, a novel \emph{hybrid} error detection framework, which combines LLM reasoning ability with the machine learning pipeline via \emph{zero-shot} prompting.
ZeroED operates in four steps, i.e., feature representation, error labeling, training data construction, and detector training.
Initially, to enhance error distinction, ZeroED generates rich data representations using LLM-driven \emph{error reason-aware} binary features, pre-trained embeddings, and statistical features.
Then, ZeroED employs LLM to \emph{holistically} label errors through in-context learning, guided by a two-step LLM reasoning process for detailed ED \emph{guidelines}.
To reduce token costs, LLMs are applied only to \emph{representative} data selected via clustering-based sampling.
High-quality training data is constructed through in-cluster label propagation and LLM augmentation with \emph{verification}.
Finally, a classifier is trained to detect all errors.
Extensive experiments on \emph{seven} datasets demonstrate that, ZeroED \emph{outperforms} state-of-the-art methods by a maximum 30\% improvement in F1 score and up to 90\% token cost reduction. 

\end{abstract} 


\section{Introduction}

\begin{figure}[t]
\centering
{\includegraphics[width=0.72\linewidth]{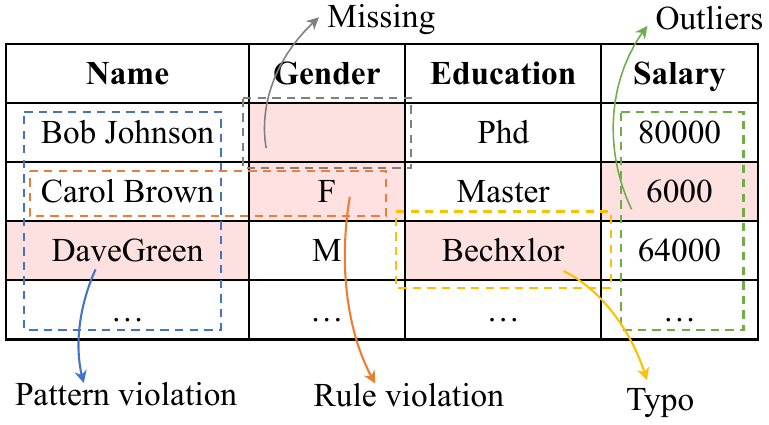}} 
\vspace{-0.04in}
\caption{Illustration of various error types in tabular data.}
\label{fig:intro_example}
\vspace{-0.18in}
\end{figure}

Dirty data have a severely negative impact on the performance of data analytical results, especially in the era of large language models (LLMs). 
When LLMs are trained on data containing errors, these mistakes can spread through their outputs, leading to false information and financial losses.~\cite{tang2023verifai, zhao2023llmsurvey,miao23imputationsurvey}.
Error detection (ED) serves as a crucial first step in data engineering by identifying quality issues before they affect downstream applications~\cite{Pang21deepanomalydetection, Ziawasch16detecting,automatic23wei}.
Real-world datasets often contain heterogeneous errors from diverse sources (e.g., typos, missing values, pattern violations)~\cite{Heidari19holodetect, Pham21spade, Huang18autodetect, raha19mahdavi}. 
Identifying these errors accurately and completely remains challenging. 
The main difficulty lies in understanding data context at various scopes, which makes manual error detection extremely time-consuming.
The example in Fig.~\ref{fig:intro_example} discusses several error types in a sampled dataset. 

\begin{example}
\label{exp:exp_intro}
    \textnormal{Fig.~\ref{fig:intro_example} states a sample of tuples from a tax dataset, containing several erroneous values highlighted in colored cells. It comprises four attributes, i.e., \emph{Name, Gender, Education, and Salary}.
    Five erroneous values are observed, representing different error types.
    The dashed boxes indicate the varying scopes of necessary \emph{data context} for ED.
    For example, missing and typo errors (i.e., NULL and `Bechxlor') can be identified by examining the \emph{individual cells}; pattern violation and outlier issues (i.e., `DaveGreen' and `6000') require pattern and domain analysis \emph{in \emph{Name} and \emph{Salary} attribute}, respectively; and rule violation problems (i.e., `F') need examining \emph{values across attributes}, where \emph{Name} determines \emph{Gender} in this case.}
\end{example}

The prior ED methods primarily rely on detecting violations of manually defined criteria, e.g., integrity constraints~\cite{Chu13holistic, Ge22mlnclean, Ebaid13nadeef}, patterns~\cite{Ebaid13nadeef}, knowledge bases~\cite{Chu15Katara, Fan12master}, and statistical thresholds~\cite{Pit2016dboost}.
They are inherently limited to detecting errors included in criteria, as shown in Table~\ref{tab:ED_comparison}.
Manually developing comprehensive criteria to capture all possible error types requires significant resources~\cite{Hao17novelcost}.
Another group of ED methods directly regards errors as values differing from clean data, typically using human labels to train a binary error detection machine learning (ML) classifier. 
As Table~\ref{tab:ED_comparison} shows, they enable holistic error detection since human experts can identify all errors.
Both ED approaches heavily rely on \emph{extensive human labor}, either through manually defined criteria or labels, thus limiting their practicability.

Moreover, for manual label-based ED methods training ML models, their \emph{learning process} suffers from two issues.
(i) \emph{Surface-level features.}
These methods extract basic features such as frequency statistics and basic lexical patterns like value frequency and TF-IDF~\cite{raha19mahdavi, Heidari19holodetect}, which lack the error semantic understanding. 
More critically, underlying error reasons are not \emph{directly} captured.
For instance, a low frequency does not necessarily indicate erroneous, and vice versa. 
(ii) \emph{Sparse and imbalanced training data.}
Real-world datasets typically have low error ratios (e.g., $<$ 40\% in CleanML~\cite{Li21cleanml} benchmark), causing class imbalance.
Limited manual labeling further reduces training samples, especially for the minority class of dirty data.
Current methods usually mitigate this issue through label propagation and error augmentation techniques~\cite{Heidari19holodetect, Pham21spade, raha19mahdavi}, which remain problematic as propagated labels lack verification, and error augmentation relies solely on string-level modifications without semantic consideration.
Besides, \emph{manual labels} are still necessary.
The emergence of LLMs has opened up new opportunities for developing ED methods with minimal human expertise via zero-shot prompting~\cite{Narayan22Foundationwrangle, wei24iterclean, fang2024llmstabular, Kojima22zeroreasoners,jia2023mill,xu2024multi}. 
They typically employ simplistic \emph{prompts}, such as querying LLM \texttt{`Is there an error in this tuple?'} on each \emph{single} tuple to detect errors.
However, these LLM prompt-based methods face two major challenges: (i) \emph{Limited ED capability}.
The way of individual tuple detection is infeasible to identify the errors that require contextual analysis among numerous tuples and attributes.
For instance, it is difficult to identify rule violations in Fig.~\ref{fig:intro_example} without knowing relationships across attributes.
(ii) \emph{Huge token consumption}.
As is well known, the computation cost of LLM is proportional to the used token number in the text processing.
Moreover, these LLM-based ED approaches have to evaluate the text of all tuples in the dataset.
Consequently, the token cost of LLM for ED would be extremely high when processing large datasets.



\begin{table}[tbp]
\caption{Comparison of existing ED methods and ZeroED. }
\vspace{-0.08in}
\begin{threeparttable}
  \centering
  \setlength{\tabcolsep}{1.1mm}
  \renewcommand{\arraystretch}{1.17}
  {\begin{tabular}{|c|c|c|c|c|c|}
  \hline
        \multirow{2.67}{*}{\makecell[c]{\textbf{Category}}} & \multirow{2.67}{*}{\centering \makecell[c]{\textbf{Methods}}} & \multicolumn{4}{c|}{\textbf{Processing errors*}} \\ \cline{3-6}
        & &
        \makecell[c]{Missing \\ \& Typos} & \makecell[c]{Pattern \\ violations} & \makecell[c]{Rule \\ violations} & \makecell[c]{Outliers} \\ \hline
        \multirow{4}{*}{\makecell[c]{Manual \\ criteria}} & dBoost~\cite{Pit2016dboost} & \ding{55} & \thincheckmark & \thincheckmark & \thincheckmark \\ 
        & Nadeef~\cite{Ebaid13nadeef} & \thincheckmark & \ding{55} & \thincheckmark & \ding{55} \\ 
        & Holistic~\cite{Chu13holistic} & \thincheckmark & \ding{55} & \thincheckmark & \ding{55} \\ 
        & Katara~\cite{Chu15Katara} & \ding{55} & \thincheckmark & \thincheckmark & \ding{55} \\ \hline
        \multirow{2}{*}{\makecell[c]{Manual \\ label}} & Raha~\cite{raha19mahdavi} & \thincheckmark & \thincheckmark & \thincheckmark & \thincheckmark \\ 
        & HoloDetect~\cite{Heidari19holodetect} & \thincheckmark & \thincheckmark & \thincheckmark & \thincheckmark \\ \hline
        \multirow{1}{*}{\makecell[c]{LLM prompt}} & FM\_ED~\cite{Narayan22Foundationwrangle} & \thincheckmark & \ding{55} & \ding{55} & \ding{55} \\ \hline
        \multirow{1}{*}{\makecell[c]{LLM label}} & ZeroED & \thincheckmark & \thincheckmark & \thincheckmark & \thincheckmark \\ 
        \hline
  \end{tabular}}
  \begin{tablenotes}
    \footnotesize
    \item[*]  Processing errors are labeled based on their experiment reports.
    \end{tablenotes}
\end{threeparttable}
\label{tab:ED_comparison}
  \vspace{-0.18in}
\end{table}

To address the limited ED capability of LLM prompt-based methods, a naive improvement is to include additional relevant tuples in the prompt as references, but this further increases token consumption.
An alternative approach is to combine LLM with manual label-based ED methods.
As shown in Table~\ref{tab:ED_comparison}, manual label-based methods holistically detect errors but require human labor. 
In contrast, LLM prompt-based methods require no manual criteria or labels but with limited ED capabilities.
This suggests a promising \emph{hybrid} approach with both strengths: \emph{with zero-shot prompting~\cite{Kojima22zeroreasoners}, leveraging LLMs to label \emph{all} error types of small data samples, then adopting an LLM-enhanced ML process to detect all errors.}
This strategy could achieve \emph{comprehensive} error detection while \emph{minimizing} both human effort and LLM token costs.

\begin{figure*}[t]
\center
  \includegraphics[width=\linewidth]{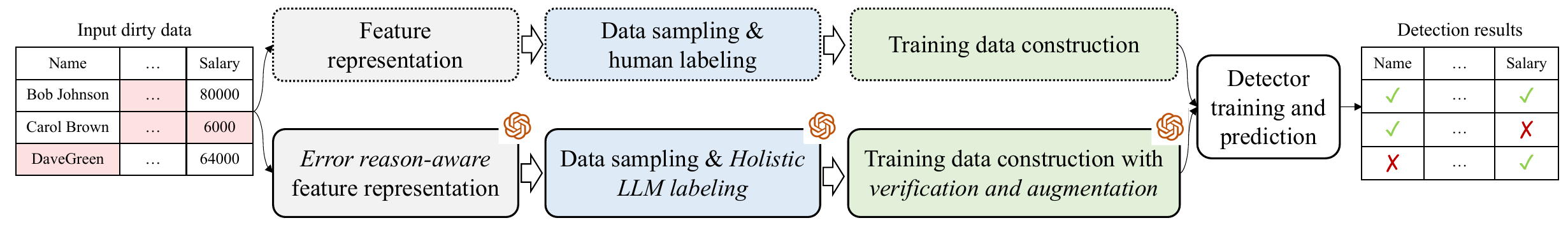}
  \caption{The comparison of our proposed hybrid framework ZeroED with previous manual label-based ones.}
 \label{fig:label_data_workflow}
 \vspace*{-0.14in}
\end{figure*}

To this end, we introduce ZeroED, a novel hybrid ED framework that combines LLMs' superior reasoning abilities with the ML process, requiring \emph{zero} pre-existing labels or criteria.
ZeroED proceeds in four main steps, i.e., feature representation, data sampling and labeling, training data construction, along with detector training and prediction.
First, ZeroED develops executable error-checking criteria, and then generates error reason-aware binary features based on data values' adherence.
Combined with pre-trained semantic embeddings and statistical features, it creates comprehensive representations.
Second, ZeroED creates error detection guidelines through a two-step process, enabling LLMs to conduct \emph{holistic} labeling through in-context learning.
To reduce token costs, ZeroED applies LLMs to label only representative data selected from a clustering-based sampling strategy.
Third, through label propagation in cluster and error augmentation, ZeroED then constructs high-quality training data with a mutual verification process.
Finally, ZeroED trains an ML classifier to detect all errors.
Our main contributions are summarized as follows.
\begin{itemize}
    \item \emph{Hybrid Zero-shot Framework.} We propose ZeroED, a novel hybrid ED framework combining \emph{holistic} LLM labeling and LLM-enhanced ML process, \emph{eliminating} the need for pre-existing labels and criteria via \emph{zero-shot} prompting.
    To the best of our knowledge, this is the \emph{first} \emph{specialized} ED framework integrating LLMs.
    \item \emph{Holistic LLM Labeling.} We improve LLMs' error detection ability via in-context learning, using comprehensive ED \emph{guidelines} derived from distribution analysis functions parsing data. This approach helps LLMs identify diverse error types while reducing \emph{token consumption}.
    \item \emph{LLM-enhanced ML process.} We leverage LLMs' reasoning abilities to optimize the machine learning process for error detection. This includes developing \emph{error reason-aware} features through executable error-checking \emph{criteria}, and
    enhancing training data quality via \emph{semantic} error augmentation and \emph{mutual verification}.
    \item \emph{Comprehensive Experiments.} Extensive experiments on \emph{seven} public datasets demonstrate that, ZeroED consistently shows superior performance, outperforming state-of-the-art methods up to 30\% in F1 score and reducing maximum 90\% token costs.
\end{itemize}

The rest of the paper proceeds as follows. We introduce the preliminaries and problem statements in Section~\ref{sec:background}. The proposed framework is elaborated and discussed in Section~\ref{sec:framework}. Comprehensive experiments and our findings are reported in Section~\ref{sec:experiments}.
Section~\ref{sec:related_work} overviews the related work. We conclude this paper in Section~\ref{sec:conclusions}.

\section{Preliminaries}\label{sec:background}
In this section, we first introduce types of errors in tabular data, and the workflow of error detection algorithms.
We then define the error detection problem in this paper.

Let $D = \{t_1, t_2, ..., t_N\}$ be a dirty tabular dataset consisting of $N$ tuples, where each $t_i$ represents an individual tuple.
The schema of $D$ is defined as $Attrs = \{a_1, a_2, ..., a_M\}$, comprising $M$ attributes. 
We use $D[i,j]$ to denote the cell value of attribute $a_j$ in tuple $t_i$. 
Let $D^*$ represent the ground truth version of dataset $D$. 
Following the existing literature~\cite{Ziawasch16detecting, Rekatsinas17holoclean, Heidari19holodetect, raha19mahdavi}, 
any value $D[i,j]$ that differs from the corresponding ground truth $D^*[i,j]$ is considered a data error.

\textbf{Types of data errors.} In tabular data, errors commonly manifest in several forms, i.e., \emph{missing values}, \emph{typos}, \emph{pattern violations}, \emph{outliers}, and \emph{rule violations}~\cite{Pham21spade, raha19mahdavi, Mahdavi20baran, Chu16datacleaning}.
Their detection requires various contexts across tuples and attributes.
Missing values are characterized by empty fields or null entries, while typos manifest as incorrect spellings or character substitutions, typically resulting from human input errors. 
These two types of errors can be simply identified through single-value examination. 
Pattern violations occur when values fail to conform to expected attribute formats, such as improper date formats or incorrectly structured email addresses. Outliers are values that significantly deviate from the dataset's statistical distribution or expected domain, potentially indicating measurement errors or genuine anomalies. 
Both pattern violations and outliers require a comprehensive understanding of the attribute's value distribution and expected patterns for detection. 
Rule violations specifically refer to inconsistencies between related attributes, such as when the capital cities are incorrect for their corresponding country values. This distinguishes them from single-attribute constraints like data patterns and domain rules, which are covered under pattern violations and outliers.

\textbf{Error detection methods.} Error detection (ED) aims to identify incorrect entries in datasets.
Most ED methods typically require external human expertise in two main forms, i.e., manual criteria and labels.
Manual criteria-based ED methods~\cite{Chu13holistic, Ebaid13nadeef, Khayyat15bigdansing, Fan08Cfds} usually utilize predefined integrity constraints (e.g., functional dependencies~\cite{Rezig21horizon, Fan08Cfds}, denial constraints~\cite{Chu13holistic,Ebaid13nadeef}), patterns (e.g., domain information, regex expressions~\cite{Huang18autodetect}), and statistical thresholds~\cite{Pit2016dboost} to identify errors based on the violations of these criteria.
These methods inherently rely on the side effects of data errors~\cite{Heidari19holodetect, Neutatz19ed2}, thus lacking comprehensiveness.

In contrast, manual label-based ED methods regard errors as values different from ground truth, adopting a machine learning process with holistic manual labeled data~\cite{Heidari19holodetect, raha19mahdavi, Pham21spade}.
As illustrated by the upper workflow in Fig.~\ref{fig:label_data_workflow}, this process begins with feature representation, where features are extracted to capture data characteristics. 
Subsequently, data samples are selected for human labeling to determine their correctness. 
These labeled instances are then used to construct training datasets, enabling the training of an ML-based detector that can identify all errors. 
Since human experts can comprehensively label all error types, they inherently can detect errors holistically. 

Recent advances in LLMs have introduced an alternative method~\cite{Narayan22Foundationwrangle,peng2025stepwise,xu2024multi,yang2024latex}, where prompt-based queries to LLMs are used to assess data correctness, reducing the reliance on traditional human expertise. But they can identify limited error types and output tokens are underutilized providing only yes/no feedback without further error reasoning insights.

\textbf{Hybrid approach.} Observing that (i) manual label-based methods, while comprehensive, demand significant human effort, and (ii) LLM-based approaches, though less human labor-intensive, typically \emph{underutilizing} LLM's reasoning capabilities, we therefore propose a framework that combines their strengths. 
As shown in Fig.~\ref{fig:label_data_workflow}, our framework follows the \emph{workflow} of manual label-based ED methods, employing the LLMs' reasoning abilities ~\cite{Huang2022TowardsRI,jia2024bridgingrelevance} to enhance the whole process via zero-shot prompting~\cite{Kojima22zeroreasoners}. 
Moreover, using the sampling strategy, LLM token costs are reduced significantly.



\textbf{Problem statement.} The error detection problem is to detect all errors in a dataset $D$ \emph{without} human-defined criteria or manual labels. 
Formally, it is a binary classification problem that assigns the most probable positive or negative class for each cell value $D[i,j]$.
A cell value is considered correctly classified if it is assigned a negative class when $D[i,j]=D^*[i,j]$, or a positive class when $D[i,j]\neq D^*[i,j]$.

To address ED as a binary classification problem \emph{without} pre-existing labels or criteria, three subproblems need to be resolved.
(i) Feature representations that can distinguish clean and erroneous data.
(ii) Holistic data labeling that covers all error types, without pre-existing labels and criteria. 
(iii) High-quality data to train an effective classifier.







\begin{figure*}[t]
\center
  \includegraphics[width=0.95\linewidth]{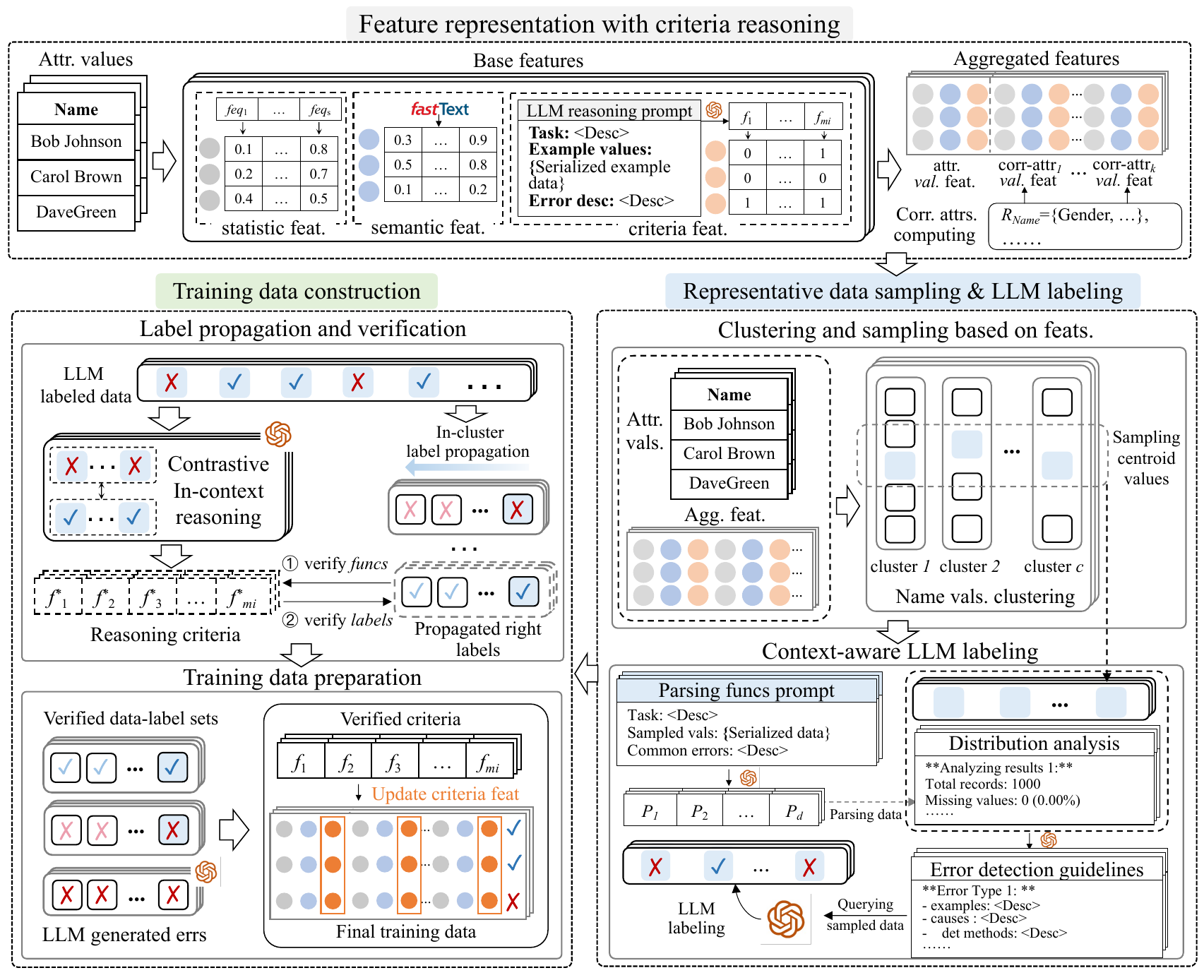}
  \caption{The detailed process of LLM-enhanced steps in ZeroED.}
 \label{fig:Framework}
 \vspace*{-0.18in}
\end{figure*}

\section{The Proposed Framework}\label{sec:framework}
In this section, we first present an overview of ZeroED. We then elaborate its four steps one by one, including LLM-driven processes of error reason-aware feature representation, data sampling and holistic LLM labeling, along with the training data construction process.
Note that, the fourth step of detector training is the same as that in previous methods.

\subsection{Framework Overview}

To achieve \emph{comprehensive} ED while \emph{minimizing} human labor and computation cost, we propose a hybrid framework combining LLM with the ML pipeline.
As shown in Fig.~\ref{fig:label_data_workflow}, ZeroED optimizes three key steps of the original workflow with LLM, i.e., feature presentation (to enable error awareness), data sampling and labeling (to minimize human efforts and computation costs), along with training data construction (to enhance training data quality). Through zero-shot prompting~\cite{Kojima22zeroreasoners}, users only need to specify simple task parameters (e.g., LLM usage budget) and universal template prompts.

Fig.~\ref{fig:Framework} details the three enhanced processes of ZeroED. 
Since error reasons and data patterns vary across attributes, ZeroED processes data by attributes. 
Initially, beyond statistical features, ZeroED also incorporates pre-trained embeddings from Fasttext~\cite{bojanowski2016enriching} to offer semantic understanding.
Then, to enhance awareness of specific error causes, LLM is applied to derive executable error-checking criteria from various perspectives.
ZeroED creates binary features based on each value's adherence to these error-checking criteria.
Combining these three features, ZeroED builds a base feature for each value.
Additionally, since tabular data values are typically correlated, ZeroED concatenates base features of \emph{related attribute} values to create the final comprehensive feature representations.

Data sampling and labeling process aim to achieve two goals: \emph{i)}  minimizing human efforts and token costs, and \emph{ii)} holistic error labeling.
To serve the \emph{first} one, ZeroED makes LLM label a few key data samples.
With the features, ZeroED adopts a \textit{clustering-based} sampling strategy.
By selecting centroid points in clusters, ZeroED ensures that the selected data are representative of both semantic and error patterns. 
The number of clusters can be decided based on user labeling budgets.
For the \emph{second} goal, ZeroED employs a two-step ED guideline generation process.
It first generates data distribution analysis functions to parse data. Then based on extracted critical statistical and contextual patterns, ZeroED creates ED guidelines that specify error examples, causes, and methods for various error types. 
With these attribute-specific guidelines, LLM can holistically grasp data relations and label errors through in-context learning.

To provide sufficient high-quality training data, ZeroED firstly \emph{expands labeled samples} by propagating LLM labels within clusters, observing that data points in the same cluster typically share similar error patterns~\cite{raha19mahdavi, Chapelle02cluster}. 
Then, ZeroED \emph{enhances data quality} through a mutual verification process, where criteria are first refined via contrastive in-context learning and then cross-checking proceeds between the propagated labeled samples and evolving criteria.
As right data typically outnumbers erroneous in real-world datasets, ZeroED further employs LLM-based error augmentation, resulting in a balanced, high-quality training dataset.

Finally, ZeroED trains a simple machine learning classifier to detect all errors, capturing complex interactions between features and data correctness.

\subsection{Feature Representation with Criteria Reasoning} 
Previous feature representation methods for error detection primarily rely on basic statistical and lexical patterns~\cite{Neutatz19ed2, Heidari19holodetect}. 
While showing relations with errors, such surface-level features inherently lack semantic understanding and error reason awareness. 
Semantic understanding is crucial for identifying semantic errors (e.g., typos and missing values) that go beyond simple statistical patterns.
Explicit reasoning about error origins can help identify context-dependent errors, like pattern and rule violations.
Therefore, ZeroED further includes pre-trained semantic embeddings and specific error-checking criteria feature through LLM reasoning. This process is fully automated, with the only human involvement being the creation of the initial prompt template for LLM reasoning.

\textbf{Statistic and semantic features.} We first introduce the applied statistics and semantic features. For statistical features, inspired by previous studies~\cite{raha19mahdavi,Huang18autodetect}, for each cell value, three types of statistic frequencies are considered, namely value frequency, vicinity frequency, and pattern frequency. 
Though limited, these different frequencies measure help identify common types of data errors.
For a cell value $D[i,j]$ in dataset $D$, value and vicinity frequency are defined as below:
\vspace*{-0.02in}
\begin{flalign*}
f_{stat}(D[i,j]) &= \frac{\{count(D[i,j]|D[i,q])\ |\ \forall a_q \in Attrs \}}{
|D[i,q]| \ \text{if } i\neq q, |D|\ \text{otherwise} }  
\end{flalign*}
where $count(D[i,j]|D[i,q])$ represents how often the value $D[i,q]$ determines $D[i,j]$. $|D[i,q]|$ refer to the occurrence number of $|D[i,q]|$ in $a_q$. When $a_q$ is equal to $a_i$, it is the value frequency, and the vicinity frequency otherwise.

For the pattern frequency, inspired by previous research for pattern violation detection~\cite{Huang18autodetect}, we generalize value $D[i,j]$ through three levels: 
L1 contains all valid characters; L2 categorizes them into letters, digits, and symbols; and L3 distinguishes upper and lower case letters.
For the generalized pattern $pat_v$, its frequency is defined as: 
\vspace*{-0.02in}
\begin{flalign*}
& f_{pat}(D[i,j]) = \frac{count(pat_v(D[i,j]))}{|pat_v(D[i,j])|} 
\end{flalign*}
where $|pat_v(D[i,j])|$ refers to its frequency in attribute $a_j$.
For example, given ``DOe123.'', L1 is ``\textmd{A[6].}'', L2 is ``\textmd{L[3]D[3]S[1]}'', L3 is ``\textmd{U[2]u[1]D[3]S[1]}'', where \textmd{A, L, U/u, D, S} denote \textmd{alphanumeric, letters, upper/lowercase, digits, symbols}, respectively. If 50/1000 values match L3, $f_{pat_3}(D[i,j]) = 0.05$.

As for the semantic embedding of cell values, instead of simple lexical features, we utilize pre-trained FastText word embeddings~\cite{joulin2017fasttext} to generate value embeddings. For cell value $D[i,j]$, ZeroED first preprocesses by tokenizing it into words and removing stop words. 
The semantic representation is then computed by averaging the embeddings of all tokens~\cite{joulin2017fasttext}. 
\begin{flalign*}
f_{sem}(D[i,j]) &= \frac{\sum_{w \in tokens(D[i,j])} Fasttext(w)}{|tokens(D[i,j])|}
\end{flalign*}
where $tokens(D[i,j])$ is the token set of the original value, and $|tokens(D[i,j])|$ represents the token number.

\textbf{Error reason-aware features.} 
Though statistical and semantic features are vital, they face challenges in distinguishing valid but uncommon data patterns and identifying errors requiring multi-tuple and multi-attribute analysis, like pattern and rule violations.
This limitation arises from their inability to comprehend the underlying error causes.
To overcome this, ZeroED makes LLMs reason about error causes, deriving error-checking criteria in executable codes (e.g., Python functions).
This approach capitalizes on LLMs' strength in code generation~\cite{zhao2023llmsurvey}, and enables richer operations beyond basic arithmetic~\cite{Ilyas15trendsdc}. 
Executing these criteria, binary features are then created based on each value's adherence, improving the distinction between erroneous and clean data.

Specifically, ZeroED derives error-checking criteria through prevailing prompt engineering with LLM~\cite{sahoo2024prompt}. 
With given prompts and data examples, LLMs can reason data-specific criteria that encode possible error reasons for 
\emph{all} possible error types, providing explicit multi-perspective validation checks.

To create prompts, the first step is \emph{tabular data serialization}, as LLMs operate fundamentally on textual contents. 
Following previous research~\cite{fang2024llmstabular, Narayan22Foundationwrangle}, for a given tuple $t_i$, we serialize it as a string of attribute-value pairs:
$serialize(t_i) = \{a_1:val_{i1}...a_M:val_{iM}\}$.
$a_j$ represents the $j$-th attribute name and $val_{ij}$ is the value of $a_j$ in $t_i$. 
In cases where an attribute value is NULL, it is represented as an empty string.

Since data patterns and correlations vary significantly in different attributes, ZeroED generates criteria by attributes. 
For each attribute $a_i$, the applied LLM $\mathcal{LM}$ is prompted with a task and role description $T_r$, common error descriptions $E$, randomly sampled tuples from dataset $D_s$ with the above serialization. This process can be formalized as:
\begin{flalign*}
& Pr_{r} = \{(T_r, serialize(D_s), E, a_i) \mid a_i \in Attrs\}\\
& \mathcal{LM}(Pr_{r}) = \{(a_i, F_i) \mid a_i \in Attrs\}
\end{flalign*}
where $Pr_{r}$ is the prompt for criteria reasoning, and the output indicates the set of criteria $F_i =\{f_1, f_2, ..., f_{ki}\}$ for each attribute $a_i$ in the dataset.

\begin{figure}[t]
\center
  \includegraphics[width=0.9\linewidth]{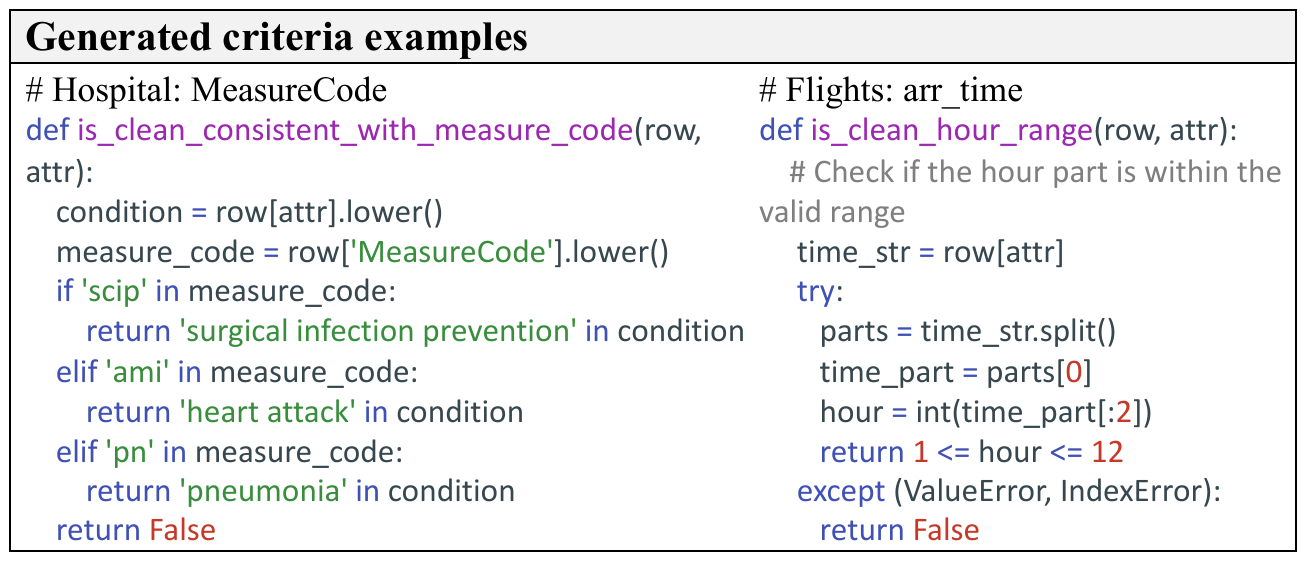}
  \caption{The examples of LLM-derived error-checking criteria.}
 \label{fig:function_example}
 \vspace*{-0.18in}
\end{figure}

As shown in Fig.~\ref{fig:function_example}, these criteria demonstrate how ZeroED uses LLM to create error-checking criteria. The Hospital criterion checks for consistency across attributes. The Flights criterion verifies if arrival times fall within a valid hour range. Both examples showcase different aspects of error reasons, meanwhile providing explainable and multi-perspective validation checks tailored to each dataset's specific characteristics.

For a given value $D[i,j]$ in attribute $a_j$, ZeroED executes each criteria $f_t$ in $F_j$ and create error-reason aware features:
\begin{flalign*}
f_{cri}(D[i,j]) = \{\text{exec}(f_t, D[i,j])\ |\ f_t \in F_j\}
\end{flalign*}
where $\text{exec}(f_k, D[i,j])$ returns a binary vector indicating whether the value satisfies the specific criterion after execution. 
This transforms the rich error-checking operations in criteria into meaningful binary feature representations.

For each value $D[i,j]$, ZeroED concatenates previous features $f_{stat}(D[i,j]), f_{pat}(D[i,j]), f_{sem}(D[i,j])$ and $f_{cri}(D[i,j])$, constructing a base feature vector $f_{base}(D[i,j])$.



\textbf{Unified Feature Representation.}
In tabular data, attributes are typically correlated, implying that the correctness of a value often depends on other related attribute values.
This makes it essential to consider features of other attribute values.
However, examining all attributes is computationally intensive and potentially counterproductive, as most attributes share minimal meaningful relationships.
This observation necessitates a strategic approach to identify strongly correlated attributes, which can provide reliable contexts for ED while maintaining computational efficiency. 

Given that the predominant relationships in tabular data are dependencies~\cite{Rezig21horizon} (e.g., Name values decide Gender ones in Fig.~\ref{fig:intro_example}), which often manifest as statistical correlations, we utilize normalized mutual information (NMI) to calculate the correlation degree between attributes.
NMI can effectively capture both linear and non-linear dependencies while providing normalized scores between 0 and 1, making it suitable for measuring correlations in tabular data.
For two attributes $a_x$ and $a_y$ in $Attr$, their mutual information is defined as:
\begin{flalign*}
& NMI(a_x, a_y) = \frac{I(a_x; a_y)}{\sqrt{H(a_x)H(a_y)}} \\
& I(a_x, a_y) = \sum_{x \in a_x} \sum_{y \in a_y} p(x,y) \log \frac{p(x,y)}{p(x)p(y)} \\
& H(a_x) = -\sum_{x \in a_x} p(x) \log p(x),\ H(a_y) = -\sum_{y \in a_y} p(y) \log p(y)
\end{flalign*}
where $p(x,y)$ is the joint probability distribution of attributes $a_x$ and $a_y$, and $p(x)$ and $p(y)$ indicates marginal probability distributions.
Since actual distributions are hard to obtain, $p(x)$ and $p(y)$ are estimated as the frequency of each value divided by total sample count, while $p(x,y)$ is the frequency of co-occurrence of values $x$ and $y$ divided by total sample count.

To identify the highly related attributes for attribute $a_i$, we compute the NMI between $a_i$ and every other attribute $a_j \in A$ in the dataset. 
ZeroED selects the attributes with the top-$k$ highest NMI scores to form the correlative attribute set $R_{a_i} = \{ a_{j_1}, a_{j_2}, \ldots, a_{j_k} \}$, providing focused contexts.


The final feature representation for cell value $D[i,j]$ combines its base features with those of its correlated attributes:
$Feat(D[i,j]) = f_{base}(D[i,j]) \oplus \{\oplus_{a_q \in R_{a_j}} f_{base}(D[i,q])\}$.
The final feature vector is within dimension $dim(f_{base}) \times (1 + k)$, where $dim(f_{base})$ is the base feature dimension and $k$ is the number of correlated attributes. 
Combining both direct and related attribute value features, our representation considers the rich contextual information for comprehensive ED.

\subsection{Representative Data Sampling and LLM Labeling}
As previously analyzed, using LLMs to directly process all data values is \emph{computationally expensive}. 
Alternatively, ZeroED only makes LLMs label a small set of key values.

\textbf{Clustering-based Data Sampling.}
Random sampling often fails to capture diverse data characteristics, especially when certain values predominate. 
This approach may also ignore minority error values, as correct data typically outnumber incorrect ones in real-world datasets. 
Instead, we propose a clustering-based sampling method based on previous feature representations.
ZeroED groups similar data into clusters and selects representative samples from each, ensuring comprehensive coverage of the dataset's diversity, including rare errors. 
Moreover, with the comparison of both normal and problematic samples, LLMs can provide more robust error labeling results in the subsequent labeling process.
Only required human effort is manually setting the number of clusters based on LLM usage budget.

ZeroED first employs the widely used $k$-means algorithm for clustering~\cite{lloyd1982kmeans}, as it naturally prioritizes denser regions of data, making it effective for sampling representative points. It is also scalable and can operate within a given clustering number.
For each attribute, the $k$-means method partitions the feature space of its values into $s$ clusters, where $s$ can be flexibly adjusted based on the LLM usage budget. 
Formally, for each attribute $a_j$ given its feature representations ${Feat(D[i,j])}^n_{i=1}$, we perform $k$-means clustering:
\begin{flalign*}
C_{a_j} & = k\text{means}(\{Feat(D[i,j])\}_{i=1}^n) \\
& = \{c_{j1}, c_{j2}, ..., c_{js}\} 
\end{flalign*}

After obtaining the clusters, ZeroED proceeds to sample the \emph{centroid} points from each cluster, which typically serves as a good representative for other data points within the cluster. 
For cluster $c_{je}$ with centroid $\mu_{je}$ we select the point $q_{c_{je}}$:
\begin{flalign*}
& q_{c_{je}} = \mathop{\arg\min}_{D[i,j] \in c_{je}} ||Feat(D[i,j]) - \mu_{je}||_2 \\
& \mu_{je} = \frac{1}{|c_{je}|} \sum_{x \in c_{je}} x
\end{flalign*}
where $|c_{je}|$ denotes the number of data points in cluster $c_{je}$, and $x$ represents individual data points within the cluster.
By sampling points near cluster centers, ZeroED ensures data characteristics coverage while maintaining efficiency. 

\textbf{Context-aware LLM Labeling.} 
As stated in Example~\ref{exp:exp_intro}, the identification of errors requires data-specific contextual analysis among tuples and attributes, which LLMs \emph{inherently lack} despite broad knowledge. 
Recent studies have demonstrated that providing explicit examples and guidelines through in-context learning significantly improves LLM performance~\cite{heetal2024annollm, zhao2023llmsurvey}. 
Inspired by this, we propose a two-step approach: first, using LLMs to generate detailed ED guidelines specifically tailored to each attribute, and then applying these guidelines to label data (correct or not). 
With these comprehensive guidelines explaining errors with specific examples, ZeroED enables LLMs to make more informed decisions.

\begin{figure}[t]
\center
  \includegraphics[width=0.91\linewidth]{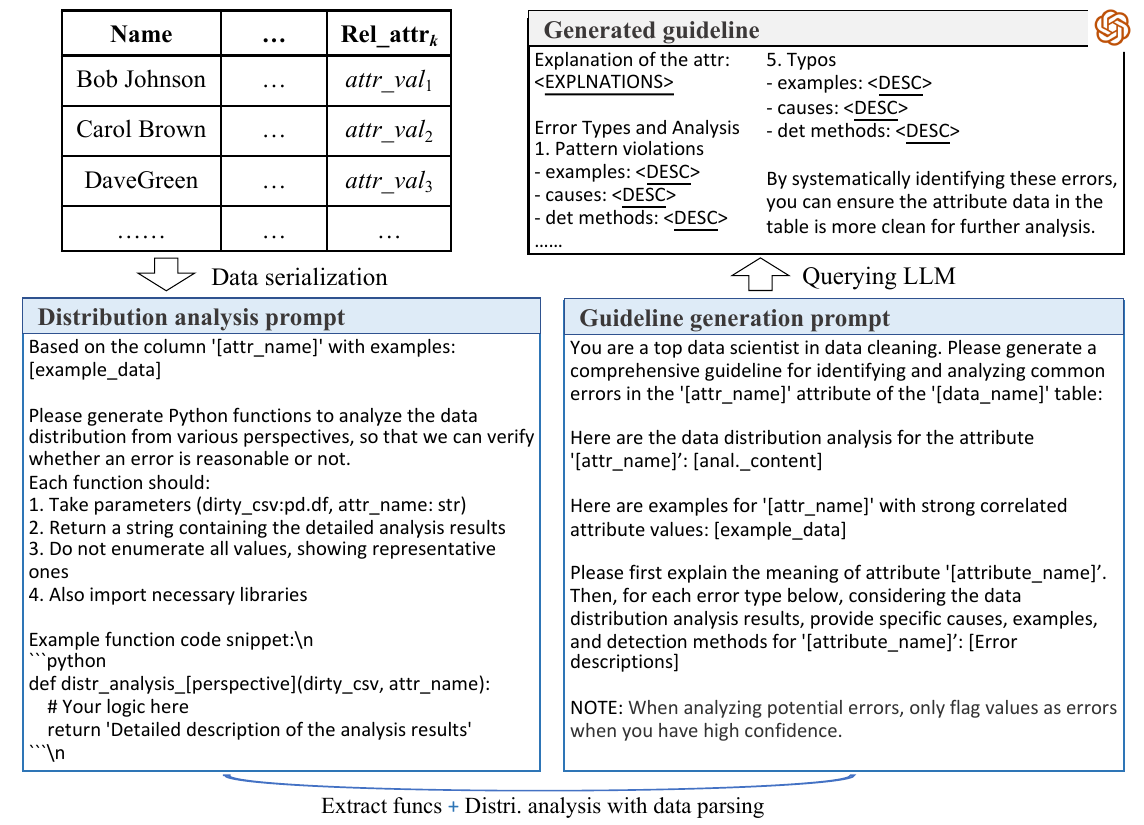}
  \caption{Guideline generation workflow and example prompts.}
 \label{fig:llm_label_prompt}
 \vspace*{-0.18in}
\end{figure}

\emph{Guideline Generation.} 
Generating beneficial data-specific ED guidelines requires an understanding of whole data distributions. 
However, directly analyzing all data examples with LLMs faces practical input length constraints~\cite{zhao2023llmsurvey}. 
Therefore, ZeroED generates error detection guidelines in two steps, as shown in Fig.~\ref{fig:llm_label_prompt}.
First, it directs LLM to create analysis functions that can parse the entire dataset and extract key (or rare) distributions. 
This enables an understanding of \emph{full-scope} data distribution without being limited by input length. 
Secondly, ZeroED combines the distribution analysis results with previously sampled representative data, making LLM generate detailed and data-specific detection guidelines.
Notably, these prompts mainly focus on task requirements, reducing the need for in-depth data knowledge. As templates, their reusability across various datasets further minimizes human effort.

This process starts by prompting LLM previous representative samples to create analysis functions.
By examining these samples, the LLM uses its inherent reasoning ability to design specific analysis functions that identify attributes' specific key features relevant for ED.
These may include common patterns, rare occurrences, and value (or pattern) distributions. 
As shown in Fig.~\ref{fig:llm_label_prompt}, the prompt mainly includes task descriptions, randomly sampled example tuples with serialization, and expected function code formats.
With functions executed to parse the whole dataset, a thorough analysis of data distribution tailored to attribute features is created.

With the distribution analysis results, ZeroED moves to generate error detection guidelines.
It is obvious that only including abstract error descriptions in the guidelines is insufficient.
For example, different attributes usually have different pattern requirements (e.g., time format and date format), and pure pattern violation descriptions are not specific enough to effectively identify errors. 
Therefore, we enhance the guideline generation process in two aspects. 
First, we provide descriptions of all common error types, i.e., missing values, typos, outliers, pattern violations, and rule violations.
These error descriptions can be generated by simply prompting LLM.
We then incorporate previously sampled representative data examples, leveraging the LLM's reasoning ability to ground the abstract error categories in instantiated, domain-specific contexts for ED.
The derived guidelines include error examples, causes, and detection methods, as shown in Fig.~\ref{fig:llm_label_prompt}.
Similarly, the guideline generation prompt consists of task descriptions, data distribution analysis results, serialized data examples, as well as error descriptions.

\emph{LLM Labeling.}
Utilizing \emph{automatically reasoned} guidelines as error judging rationale, ZeroED can then \emph{comprehensively} label data with \emph{zero-shot} prompting~\cite{Kojima22zeroreasoners}.
Leveraging LLMs' in-context learning capabilities and pre-trained knowledge, this \emph{labeling} process eliminates the need for pre-existing data-specific labels (and criteria), parameter-tuned training, and manual annotation.
Limited human intervention is only required to specify high-level task and workflow design, which are universal across datasets.

Specifically, for each attribute, we present LLM with both corresponding detection guidelines and selected samples for analysis. 
LLM examines each value by comparing it against the guidelines, enabling the comprehensive detection processes.
To handle large-scale datasets efficiently, a batch processing approach is applied. 
Rather than processing the sampled data at once, we divide it into manageable batches to ensure LLM can focus on a reasonable number of instances at a time, leading to more reliable error detection.
The labeling process can be expressed as:
\begin{flalign*}
& Pr_l = \{(T_{l}, G_i, B_w, a_i) \mid a_i \in Attrs\} \\
& \mathcal{LM}(Pr_l) = \{(v_{wj},l_{wj})\ |\ l_{wj} \in \{0, 1\}, j\in \{1,...,|B_w|\}\}
\end{flalign*}
where $T_l$ is the task descriptions, $G_i$ represents the task description and guideline for attribute $a_i$,
$B_w$ denotes the $w$-th batch of data instances, $v_{wj}$ represents the $j$-th value in batch $w$, and $l_{wj}$ indicates the label (i.e., error or not) assigned to value $v_{wj}$.
LLM examines each value against the guidelines $G_i$ to identify potential errors.

Notably, in each batch, ZeroED integrates not only the values in $a_i$ but also values of correlated attributes in $R_{a_i}$. 
Through this context-aware approach, we can enable comprehensive error detection while maintaining efficiency.
\subsection{Constructing Training Data}
After completing the LLM labeling process, our goal is to train simple detectors to detect remaining errors. 
However, this task presents two challenges. 
First, \emph{limited} LLM labeled data can lead to detectors overfitting. 
Second, real-world datasets are typically \emph{class-imbalanced}, containing far more correct values than errors. 
These challenges call for a robust approach to construct sufficient and high-quality training data for building reliable error detectors.

As shown in Algorithm~\ref{algo:train_data}, firstly, ZeroED propagates the LLM labels within clusters to expand the training data pool (Line 1), observing that data within the same cluster are likely to share the same class label~\cite{Chapelle02cluster, raha19mahdavi},
However, this propagation may introduce unreliable labels.
A direct way is to use the previous criteria set $\{F_i\}^M_{i=1}$ to verify these labels. 
Yet, this is not ideal since these criteria are derived from random samples, and may consequently miss certain error patterns.

We therefore propose \emph{contrastive in-context prompting} to enhance criteria quality (Lines 4-7). 
This strategy works by structuring prompts to present both clean and erroneous value groups, guiding LLM to recognize subtle distinctions between correct and incorrect data. 
Through such targeted prompting, $\mathcal{LM}$ develops a more refined understanding of both error patterns and validity requirements within specific data contexts, resulting in enhanced error-checking criteria $F^*$.

\begin{algorithm}[t]
\small
\caption{Training Data Construction}
\DontPrintSemicolon
\LinesNumbered
\SetNlSty{footnotesize}{}{:}
\label{algo:train_data}
\KwIn{Clusters $C$, LLM-labeled data $L$, Attributes $Attrs$, and LLM $\mathcal{LM}$}
\KwOut{Training dataset $T$, updated criteria $F^*$}

$P \leftarrow \text{PropagateLabels}(C, L)$\;
$F^* \leftarrow \emptyset$\;
\For{$a_i \in Attrs$}{
    $V_{it} \leftarrow \text{SampleErrorValues}(L, a_i)$\;
    $V_{if} \leftarrow \text{SampleRightValues}(L, a_i)$\;
    $Pr_c \leftarrow \text{ConstructContrastivePrompt}(V_{it}, V_{if})$\;
    $F_i^* \leftarrow \mathcal{LM}(Pr_c)$\;
    /* Verify criteria with right labels*/\;
    $P^i_{right} \leftarrow \text{FilterLabelsRight}(P^i)$\;
    \For{$f^*_i \in F^*_i$} {
        \If{$\text{AccOnDataRight}(f^*_i, P^i_{right}) < 0.5$}{
            $F^*_i.remove(f^*_i)$\;
        }
    }
    
/* Verify data with reliable criteria*/ \;
    \For{$data \in P^i_{right}$}{
        \If{$\text{PassRateOnCriteria}(data, F^*_i) < 0.5$}{
            $P^i_{right}.remove(data)$\;
        }
    }
    $P_{right}.update(P^i_{right})$\;
    $F^*.add(F_i^*)$\;
}
$P.update(P_{right})$\;
$GenErrs \leftarrow \text{GenerateErrs}(P, \mathcal{LM})$\;
$T \leftarrow \text{CombineData}(P, GenErrs)$\;
\Return{$T$, $F^*$}
\end{algorithm}

To ensure the quality of both training data and criteria, ZeroED implements a \emph{mutual verification} process between the enhanced criteria and propagated labels.
Observing that correct values often make up more than \emph{0.5} of real-world datasets, we use data with correct labels to verify our criteria, applying a 0.5 accuracy threshold to identify data with correct labels (Lines 8-14).
This process helps discard unreliable criteria.
ZeroED then uses these verified criteria to examine propagated correct labels, removing data where over 50\% of criteria indicate incorrectness (Lines 15-20).
This mutual verification helps enhance the quality of both our criteria and training data.

To address the challenge of class imbalance, we leverage LLM to create additional error examples. (Lines 24-25). By analyzing error reasons in verified labeled data, LLMs create more error instances that maintain semantic similarity while reflecting realistic scenarios, effectively augmenting the minority error class while preserving data quality.
The final training dataset combines the verified propagated labels and synthetic error examples (Line 26). This comprehensive approach yields a balanced, high-quality training dataset that captures diverse error patterns while maintaining semantic validity with only human effort involved in prompt formulation with defined configurations, enabling the training of a robust ED classifier.
Within high-quality training data, ZeroED trains a simple Multilayer Perceptron (MLP) model to classify all cell values as clean or dirty, which captures the non-linear interactions between feature representations and value correctness.



The commonly used cross-entropy loss function is employed as the objective function for model optimization, which is defined as follows.
\begin{flalign*}
    \mathcal{L}(y'_v, \hat{y}_v)=-\frac1{|B_w|}\sum_{v\in B_w}[y'_v\log\hat{y}_v+(1-y'_v)\log(1-\hat{y}_v)]
\end{flalign*}
where $\hat{y}_v$ represents the prediction of MLP for value $v$ , $y'_v$ denotes the label in batch $B_w$, and $|B_w|$ is size of the batch.
Upon completion of training, the MLP classifier is applied to predict the correctness of all values in the target dataset $D$. 

\subsection{Discussions}
ZeroED presents a novel hybrid resolution for zero-shot tabular data error detection, with the combination of LLM labeling and LLM-enhanced ML process.
Compared with traditional non-LLM methods, ZeroED \emph{eliminates} the need for manual criteria or labeling.
Moreover, harnessing LLMs' vast background knowledge, ZeroED can better understand data patterns and relationships. This allows for processing a broader range of error types and adapting to diverse error scenarios.

In contrast to other LLM-based ED methods, instead of direct error detection with individual tuples, ZeroED employs LLMs' reasoning abilities to derive error-checking criteria and error detection guidelines.
This makes it can \emph{holistically} identify errors like human experts.
Moreover, ZeroED improves the machine learning process through better error-distinguishable feature representations and higher-quality training data, leading to competitive performance while keeping low token costs.
Experimental results demonstrate ZeroED's effectiveness with up to 30\% improvement in F1 score while reducing token costs by up to 90\%. 
This makes ZeroED a practical solution, marking an important advance in fully automated error detection.

\begin{table}[tbp]
\small
\caption{The information of the evaluation datasets.}
\label{tab:DatasetDetailed}
  \centering
  \setlength{\tabcolsep}{0.2mm}
  \renewcommand{\arraystretch}{1.2}
  \begin{tabular}{|l|r|r|r|r|r|r|r|r|}
  \hline
        \textbf{Name}      & \makecell[c]{\textbf{\#Tuples}}  & \makecell[c]{\textbf{\#A.}} & \makecell[c]{\textbf{Err.(\%)}} & \makecell[c]{\textbf{MV(\%)}} & \makecell[c]{\textbf{PV(\%)}} & \makecell[c]{\textbf{T(\%)}} & \makecell[c]{\textbf{O(\%)}} & \makecell[c]{\textbf{RV(\%)}}\\ 
        \hline
    \textbf{Hospital} & 1,000 & 20 & 4.82 & 0 & 2.75 & 2.71 & 2.98 & 2.05 \\ \hline
    \textbf{Flights} & 2,376 & 7 & 34.51 & 16.22 & 20.12 & 13.92 & 17.52 & 34.51 \\ \hline
    \textbf{Beers} & 2,410 & 11 & 12.98 & 0.90 & 9.14 & 2.43 & 1.09 & 1.12 \\ \hline
    \textbf{Rayyan} & 1,000 & 11 & 29.19 & 15.31 & 9.42 & 3.23 & 8.47 & 11.40 \\ \hline
    \textbf{Billionaire} & 2,615 & 22 & 9.84 & 2.41 & 3.14 & 1.35 & 3.80 & 0.56 \\  \hline
    \textbf{Movies} & 7,390 & 17 & 4.97 & 2.22 & 2.32 & 0.03 & 2.64 & 0 \\ \hline
    \textbf{Tax} & 200,000 & 22 & 0.11 & 0.01 & 3.36 & 0.04 & 0.08 & 0.03 \\ 
    \hline
  \end{tabular}
  \vspace{-0.18in}
\end{table}

\section{Experiments}\label{sec:experiments}
We evaluate the performance of ZeroED against six state-of-the-art ED methods. Experiments are conducted on a server with an Intel Xeon Gold 6326 CPU (2.90GHz), 3*A40 GPUs, 512GB RAM, running Ubuntu 20.04.6 LTS. The source code is available at https://github.com/ZJU-Data-Governance-and-Services-Team/ZeroED.

\begin{table*}[t]
\small
\caption{Performance comparison of error detection methods, with the best results highlighted in bold. The improvements in F1 score over all baselines are statistically significant (i.e., paired t-tests with $p<0.05$).}
  \label{tab:Effectiveness}
  \renewcommand{\arraystretch}{1.1}
  \setlength{\tabcolsep}{0.95mm}
  \centering
    \begin{tabular}{|c|ccc|ccc|ccc|ccc|ccc|ccc|}
    \hline
    \multicolumn{1}{|c|}{\multirow{2}{*}{\textbf{Methods}}} & \multicolumn{3}{c|}{\textbf{Hospital}} & \multicolumn{3}{c|}{\textbf{Flights}} & \multicolumn{3}{c|}{\textbf{Beers}} & 
    \multicolumn{3}{c|}{\textbf{Rayyan}} & \multicolumn{3}{c|}{\textbf{Billionaire}} & \multicolumn{3}{c|}{\textbf{Movies}} \\ \cline{2-19}
          & Prec & Rec & F1 & Prec & Rec & F1 & Prec & Rec & F1 & Prec & Rec & F1 & Prec & Rec & F1 & Prec & Rec & F1 \\
    \hline
     dBoost & 0.887 & 0.355 & 0.507 & 0.753 & 0.582 & 0.657 & 0.535 & \textbf{0.997} & 0.697 & 0.515 & 0.414 & 0.459 & \textbf{0.795} & 0.497 & 0.612 & 0.555 & 0.412 & 0.473\\ 
     Nadeef & 0.061 & 0.257 & 0.059 & 0.420 & 0.927 & 0.578 & 0.135 & 0.089 & 0.107 & 0.742 & 0.556 & 0.632 & 0.145 & 0.083 & 0.106 & \textbf{1.000} & 0.104 & 0.189 \\ 
     Katara & 0.439 & 0.071 & 0.122 & 0 & 0 & 0 & 0 & 0 & 0 & 0 & 0 & 0 & 0.101 & 0.013 & 0.022 & 0 & 0 & 0 \\ \hline 
     ActiveClean & 0.049 & 0.088 & 0.074 & 0.350 & \textbf{0.959} & 0.513 & 0.130 & 0.996 & 0.230 & 0.292 & \textbf{1.000} & 0.452 & 0.098 & \textbf{0.935} & 0.179 & 0.109 & 0.006 & 0.011 \\ 
     Raha & 0.727 & 0.068 & 0.125 & 0.719 & 0.612 & 0.591 & 0.742 & 0.636 & 0.685 & 0.532 & 0.350 & 0.422 & 0.278 & 0.126 & 0.174 & 0.376 & 0.371 & 0.373 \\ \hline 
     FM\_ED & 0.665 & 0.638 & 0.651 & 0.926 & 0.513 & 0.660 & 0.866 & 0.076 & 0.139 & \textbf{0.793} & 0.568 & 0.662 & 0.628 & 0.727 & 0.674 & 0.793 & 0.461 & 0.583\\ \hline 
     ZeroED & \textbf{0.936} & \textbf{0.715} & \textbf{0.811} & \textbf{0.935} & 0.586 & \textbf{0.722} & \textbf{0.888} & 0.689 & \textbf{0.774} & 0.778 & 0.692 & \textbf{0.732} & 0.768 & 0.767 & \textbf{0.767} & 0.724 & \textbf{0.812} & \textbf{0.765} \\ 
    \hline
    \end{tabular}
    \vspace*{-0.06in}
\end{table*}

\subsection{Experiment Settings}
\textbf{Datasets.} 
Our experiments utilize seven datasets, encompassing both real-world and synthesized errors. Five datasets, i.e., \emph{Hospital}, \emph{Flights}, \emph{Beers}, \emph{Movies}, and \emph{Rayyan}, contain real-world errors, while \emph{Billionaire} and \emph{Tax} feature manually injected errors. 
We selected these datasets for their representation of common, realistic error types, i.e., outliers (O), typographical errors (T), pattern violations (PV), missing values (MV), and rule violations (RV). 
Table~\ref{tab:DatasetDetailed} presents comprehensive details about these datasets.
We report overall error rate (Err.) and individual error rates by type. As no explicit methods exist for type detection, we classify them as follows: T include errors within edit distance $\leq$3 from clean data;
PV represent error formats unseen in the clean data; MV include explicit and implicit placeholders; RV cover rule violations; O are errors with $<$1\% frequency.
\emph{Hospital} and \emph{Flights} are well-established ED benchmarks~\cite{raha19mahdavi,Neutatz19ed2,Pham21spade,Rekatsinas17holoclean}. 
\emph{Beers}~\cite{databeers} dataset is cleaned manually, whereas \emph{Movies} is sourced from the Magellan repository~\cite{das2015magellan}, cleaned with duplicate tuple labels. 
\emph{Rayyan} is another real-world dataset, cleaned by its original owners.
To expand the range of error types, we also create a dirty \emph{Billionaire}~\cite{databillion} dataset.  Using publicly available code~\cite{errorgurl}, we introduced all five types of errors. 
Tax is a large synthetic dataset from the BART repository~\cite{Arocena15bart}. We use this dataset for scalability evaluations.

\textbf{Baselines.}
We evaluate ZeroED against diverse state-of-the-art ED methods for tabular data. Our comparison includes methods requiring external manual criteria: dBoost~\cite{Pit2016dboost} with statistical configuration, Nadeef~\cite{Ebaid13nadeef} using pre-defined constraints and patterns, and Katara~\cite{Chu15Katara} leveraging knowledge bases. We also include state-of-the-art manual label based approaches Raha~\cite{raha19mahdavi}. 
For comprehensive comparison, we include ActiveClean~\cite{Krishnan16activeclean}, which employs downstream models for error detection.
Additionally, we incorporate FM\_ED~\cite{Narayan22Foundationwrangle} as a representative of LLM-driven methods
This comparison spans various ED paradigms and techniques, enabling a holistic evaluation of ZeroED's capabilities.


\textbf{Evaluation Metrics}.
To comprehensively evaluate ED performance, we employ precision (Prec), recall (Rec), and F1-score (F1) metrics, which are widely adopted in previous literature~\cite{Rekatsinas17holoclean,Abdelaal23rein,Ziawasch16detecting,raha19mahdavi,Heidari19holodetect}
Precision is the percentage of correctly identified errors among all identified errors. Recall is the proportion of identified errors out of all actual errors. F1-score is the harmonic mean of precision and recall. Higher values across them indicate better ED performance.


\begin{figure}[t]
\centering
\includegraphics[width=0.78\linewidth]{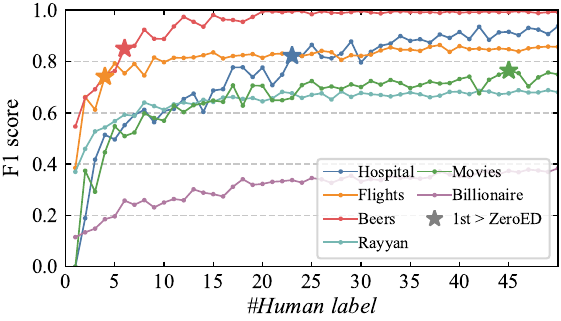}
\vspace{-0.1in}
\caption{Raha performance via active learning.}
\label{fig:raha_label_num}
\vspace*{-0.2in}
\end{figure}

\textbf{Implementation Details.}
All open-source LLMs are run on vllm~\cite{kwon24vllm}, including Qwen~\cite{yang2024qwen2technicalreport} and Llama~\cite{dubey2024llama3herdmodels} series.
The closed-source gpt-4o-mini-2024-05-13 model through API calls is also incorporated to assess runtime and token costs. 
For reproducibility, LLM query parameters are set with a temperature of 0 and a maximum output of 4096 tokens.
By default, we use Qwen2.5-72b to label 5\% of data, with the clustering number determined by multiplying data size. 
2 related attributes are used except in hyperparameter experiments.
The criteria are generated within two rounds of prompting.
Our MLP model comprises two layers with ReLU activation functions. To address the input length constraints of LLMs, we segmented the data into sequential batches of 20 tuples each for prompts requiring data input. 
For baselines requiring integrity constraints, patterns, and knowledge bases, we incorporated these from existing public code~\cite{raha19mahdavi, Pham21spade}.
To meet the scenarios requiring minimal human efforts, we used 2 labeled tuples per dataset for ED methods requiring manual labels. 
Notably, ZeroED \emph{does not depend} on manually defined constraints, patterns, knowledge bases, or labels. 
Each reported result represents the average of three repeated experiments. Other parameters are set to their default values.


\begin{table*}[t]
\small
\caption{Ablation study of ZeroED. 
All decreases in F1 score values with process removal are statistically significant (i.e., two-sided t-tests with $p<0.05$).}
  \label{tab:ablation}
  \renewcommand{\arraystretch}{1.15}
  \setlength{\tabcolsep}{0.9mm}
  \centering
    \begin{tabular}{|c|ccc|ccc|ccc|ccc|ccc|ccc|}
    \hline
    \multicolumn{1}{|c|}{\multirow{2}{*}{\makecell[c]{\textbf{Ablation} \\ \textbf{Process}} }} & \multicolumn{3}{c|}{\textbf{Hospital}} & \multicolumn{3}{c|}{\textbf{Flights}} & \multicolumn{3}{c|}{\textbf{Beers}} & 
    \multicolumn{3}{c|}{\textbf{Rayyan}} & \multicolumn{3}{c|}{\textbf{Billionaire}} & \multicolumn{3}{c|}{\textbf{Movies}} \\ \cline{2-19}
          & Prec & Rec & F1 & Prec & Rec & F1 & Prec & Rec & F1 & Prec & Rec & F1 & Prec & Rec & F1 & Prec & Rec & F1 \\
    \hline
     \textit{w/o.} Guid. & 0.926 & 0.697 & 0.795 & 0.905 & 0.577 & 0.705 & 0.852 & 0.370 & 0.516 & 0.751 & 0.618 & 0.678 & 0.497 & 0.752 & 0.598 & 0.662 & 0.805 & 0.727 \\ \hline
     \textit{w/o.} Crit. & 0.652 & 0.573 & 0.609 & \textbf{0.965} & 0.534 & 0.688 & 0.923 & 0.581 & 0.714 & 0.697 & 0.611 & 0.651 & 0.624 & 0.718 & 0.667 & 0.543 & 0.664 & 0.598 \\ \hline 
     \textit{w/o.} Corr.& 0.903 & 0.692 & 0.784 & 0.681 & 0.573 & 0.623 & \textbf{0.931} & 0.395 & 0.555 & 0.696 & 0.457 & 0.552 & 0.685 & 0.702 & 0.693 & 0.392 & 0.738 & 0.511\\ \hline 
     \textit{w/o.} Veri. & 0.932 & 0.664 & 0.775 & 0.933 & 0.571 & 0.708 & 0.890 & 0.682 & 0.772 & 0.774 & 0.663 & 0.714 & 0.768 & 0.767 & 0.767  & 0.724 & 0.812 & 0.765\\ \hline 
     ZeroED & \textbf{0.936} & \textbf{0.715} & \textbf{0.811} & 0.935 & \textbf{0.586} & \textbf{0.722} & 0.888 & \textbf{0.689} & \textbf{0.774} & \textbf{0.778} & \textbf{0.692} & \textbf{0.732} & \textbf{0.768} & \textbf{0.767} & \textbf{0.767} & \textbf{0.72}4 & \textbf{0.812} & \textbf{0.765} \\ 
    \hline
    \end{tabular}
    \vspace{-0.18in}
\end{table*}

\subsection{Comparison Study}
We first comprehensively explore the error detection performance of different error detection methods, across six datasets, shown in Table~\ref{tab:Effectiveness}
For justification, we also evaluate the performance of Raha with different numbers of labeled tuples, as illustrated in Fig.~\ref{fig:raha_label_num}. 
ZeroED achieves superior performance in most scenarios, particularly excelling in precision and F1 score.
When compared to traditional criteria-based error detection methods such as dBoost, Nadeef, and Katara, ZeroED exhibits superior performance across almost all metrics. This is primarily due to the fact that criteria violations do not always indicate absolute errors, which can lead to suboptimal ED performance.
Katara shows minimal detection results on Flights, Beers, and Rayyan due to the absence of relevant knowledge bases.
Nadeef achieves perfect precision on Movies, due to the limited but precision pattern criteria.
ZeroED also demonstrates significant improvements over Raha. As shown in Fig.~\ref{fig:raha_label_num}, the star points indicate the first time Raha outperforms ZeroED. Notably, except for Flights and Beers, Raha requires over 20 labeled tuples to achieve performance comparable to ZeroED. This underscores ZeroED's effectiveness without the need for labeled data.
ActiveClean struggles to differentiate between errors and clean data in the Flights and Rayyan datasets due to its simple feature extraction method, leading it to treat all data as incorrect.
ZeroED's outperforming of FM\_ED demonstrates that its novel framework, which combines LLMs with an ML pipeline, offers superior accuracy and comprehensiveness compared to purely LLM-based detection approaches.

\begin{figure}[t]
\centering
\includegraphics[width=\linewidth]{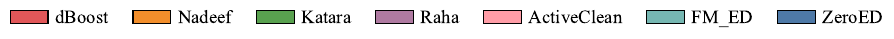}
\vspace{-0.2in}
\\
\hspace{-0.08in}\subfigure[\emph{Runtime across datasets}]{\raisebox{-0.2cm}{\includegraphics[height=0.34\linewidth]{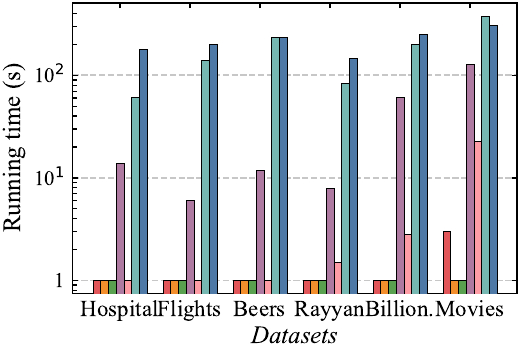}}}
\subfigure[\emph{Runtime with various data sizes}]{\raisebox{-0.2cm}
{\includegraphics[height=0.34\linewidth]{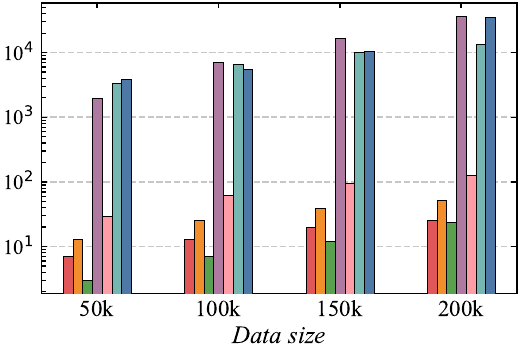}}}
\vspace{-0.08in}
\caption{Running time eval. across datasets and data sizes.}
\label{fig:runtime}
\vspace*{-0.18in}
\end{figure}

\subsection{Running Time and Token Consumption Evaluation}
\subsubsection{Running time} We analyzed the end-to-end runtime across datasets, with a special focus on different-sized subsets of the largest Tax dataset. As shown in Fig.~\ref{fig:runtime}, traditional methods like dBoost, Nadeef, and Katara execute quickly due to simple heuristics. FM\_ED and ZeroED run slower due to API limitations. In the largest Tax datasets, ZeroED's runtime becomes comparable to Raha. 
Moreover, rapidly improving LLM processing speeds (e.g., GPT-4o is 4x faster than original GPT-4) suggest the runtime will continue to decrease.

\subsubsection{Token consumption} 
\begin{figure}[t]
\centering
\includegraphics[width=0.92\linewidth]{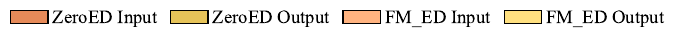}
\vspace{-0.06in}
\\
\hspace{-0.04in}\subfigure[\emph{Token cost across datasets}]{\raisebox{-0.2cm}{
\includegraphics[height=0.33\linewidth]{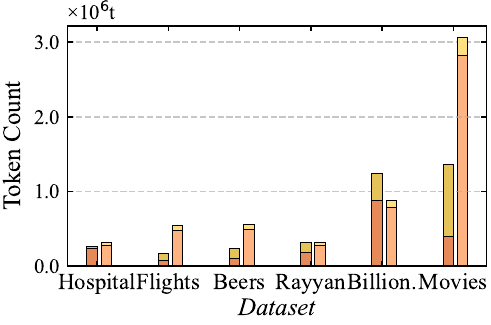}}}
\hspace{-0.06in}\subfigure[\emph{Token cost within data sizes}]{\raisebox{-0.2cm}{
\includegraphics[height=0.33\linewidth]{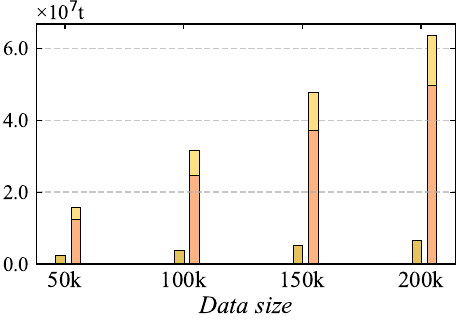}}}
\vspace{-0.08in}
\caption{Token consumption eval. across datasets and data sizes.}
\label{fig:token}
\vspace{-0.2in}
\end{figure}

We compare token costs between FM\_ED and ZeroED. Fig.~\ref{fig:token} shows that 
ZeroED generally uses fewer tokens as dataset size grows, except for Billionaire dataset due to its high attribute count and small data size. Notably, FM\_ED typically uses more input tokens, while ZeroED focuses on output tokens. This suggests ZeroED more effectively leverages the LLM for insights. Experiments on the Tax dataset demonstrate ZeroED's superior scalability, maintaining modest token growth compared to FM\_ED's steep increase, achieving over \emph{90\%} token cost reduction at the maximum dataset size.

\begin{table*}[t]
\vspace*{-0.1in}
\small
\caption{Detection performance comparison of ZeroED with different LLMs}
  \label{tab:llm}
  \renewcommand{\arraystretch}{1.15}
  \setlength{\tabcolsep}{0.9mm}
  \centering
    \begin{tabular}{|l|ccc|ccc|ccc|ccc|ccc|ccc|}
    \hline
    \multicolumn{1}{|c|}{\multirow{2}{*}{\textbf{LLMs}}} & \multicolumn{3}{c|}{\textbf{Hospital}} & \multicolumn{3}{c|}{\textbf{Flights}} & \multicolumn{3}{c|}{\textbf{Beers}} & 
    \multicolumn{3}{c|}{\textbf{Rayyan}} & \multicolumn{3}{c|}{\textbf{Billionaire}} & \multicolumn{3}{c|}{\textbf{Movies}} \\ \cline{2-19}
          & Prec & Rec & F1 & Prec & Rec & F1 & Prec & Rec & F1 & Prec & Rec & F1 & Prec & Rec & F1 & Prec & Rec & F1 \\
    \hline
     
     GPT-4o-mini & 0.164 & 0.691 & 0.265 & 0.687 & 0.492 & 0.574 & 0.532 & 0.617 & 0.571 & 0.447 & 0.545 & 0.491 & 0.311 & 0.501 & 0.384 & 0.208 & 0.736 & 0.325\\ \hline 
     Llama3.1-8b & 0.678 & \textbf{0.815} & 0.755 & 0.820 & \textbf{0.594} & 0.689 & 0.629 & 0.434 & 0.514 & \textbf{0.798} & 0.627 & 0.702 & \textbf{0.819} & 0.702 & 0.756 & 0.484 & 0.782 & 0.598 \\ \hline 
     Llama3.1-70b & 0.687 & 0.763 & 0.723 & 0.886 & 0.583 & 0.703 & \textbf{0.907} & 0.502 & 0.647 & 0.774 & 0.682 & 0.724 & 0.612 & 0.706 & 0.656 & 0.686 & 0.735 & 0.710 \\ \hline 
     Qwen2.5-7b & 0.532 & 0.776 & 0.631 & 0.802 & 0.552 & 0.654 & 0.712 & 0.543 & 0.616 & 0.779 & 0.596 & 0.675 & 0.418 & 0.318 & 0.361 & 0.300 & 0.449 & 0.360 \\ \hline
     Qwen2.5-72b & \textbf{0.936} & 0.715 & \textbf{0.811} & \textbf{0.935} & 0.586 & \textbf{0.722} & 0.888 & \textbf{0.689} & \textbf{0.774} & 0.778 & \textbf{0.692} & \textbf{0.732} & 0.768 & \textbf{0.767} & \textbf{0.767} & \textbf{0.724} & \textbf{0.812} & \textbf{0.765} \\ 
    \hline
    \end{tabular}
    \vspace{-0.1in}
\end{table*}

\begin{figure*}[t]
\centering
\subfigure[Hospital]{\raisebox{-0.1cm}{\includegraphics[width=0.3\linewidth]{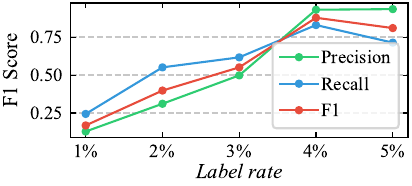}}}
\subfigure[Flights]{\raisebox{-0.1cm}{\includegraphics[width=0.3\linewidth]{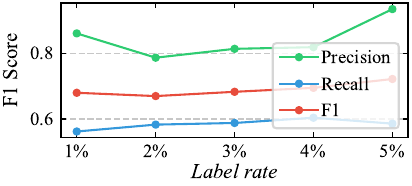}}}
\subfigure[Beers]{\raisebox{-0.1cm}{\includegraphics[width=0.3\linewidth]{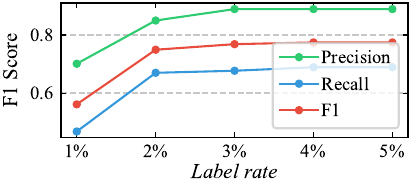}}}
\vspace*{-0.08in}
\\
\subfigure[Rayyan]{\raisebox{-0.1cm}{\includegraphics[width=0.3\linewidth]{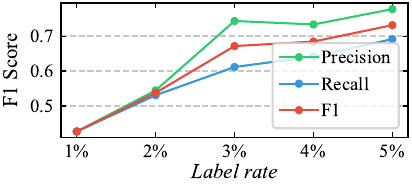}}}
\subfigure[Billionaire]{\raisebox{-0.1cm}{\includegraphics[width=0.3\linewidth]{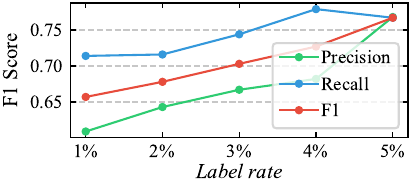}}}
\subfigure[Movies]{\raisebox{-0.1cm}{\includegraphics[width=0.3\linewidth]{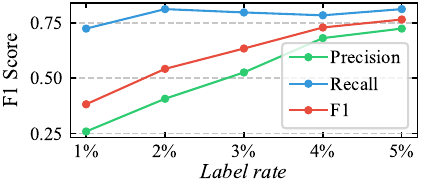}}}
\vspace*{-0.1in}
\caption{Error detection performance under different data label rates (clustering number) using LLM.}
\label{fig:label_rate}
\vspace{-0.16in}
\end{figure*}

\subsection{Ablation Study} 
We conducted ablation studies on ZeroED by removing individual components, i.e., guideline generation (Guid.), criteria reasoning (Crit.), correlated attributes calculation (Corr.), and training data verification and error generation (Veri.) processes. 
Table~\ref{tab:ablation} shows that removing any component from ZeroED leads to a decrease in performance, with all decreases in F1 score being statistically significant ($p<0.05$). 
The results highlight that guideline generation, criteria reasoning, and correlated attribute calculation are particularly crucial for ZeroED. 
For Hospital and Flights, which contain easily detectable errors such as simple typos and missing values, removing guidelines has minimal effect. 
However, for complex datasets, with a broader range of errors, the guidelines prove critical.
As for training data verification and error generation process, it is also crucial for Hospital, Flights, and Beers, but has minimal impact on other datasets. 
This is possibly because other datasets have relatively more diverse errors,
While removing criteria reasoning and correlated attributes calculation may slightly increase precision, this marginal improvement did not offset the overall drop in model robustness.
\subsection{Parameter Evaluation}
\subsubsection{Effect of LLM label rate} \normalfont(Clustering number)
Fig.~\ref{fig:label_rate} illustrates ZeroED's performance as the LLM label rate increases from 1\% to 5\%. 
The clustering number for each dataset is calculated as $data\_size * label\_rate$. 
We observe general metric improvements with higher label rates. Precision consistently improves across most datasets, while recall shows less consistency, with slight decreases possibly indicating overfitting or noise for some datasets. F1 scores improve steadily, confirming that higher label rates enhance overall performance, though at varying rates. These results underscore ZeroED's effectiveness in leveraging increased labeled data.
\subsubsection{Effect of LLMs}
Table~\ref{tab:llm} presents the F1 scores of ZeroED using various open-source and closed-source LLMs, including the Qwen series~\cite{yang2024qwen2technicalreport}, Llama series~\cite{dubey2024llama3herdmodels}, and GPT-4o-mini.
The results reveal significant variability in performance. Qwen2.5-72b consistently achieved the highest F1 scores across most datasets, demonstrating superior performance. In contrast, GPT-4o-mini underperformed, highlighting its limitations in handling diverse error types.
Larger models, such as Llama3.1-70b and Qwen2.5-72b, generally outperformed smaller ones, indicating their enhanced ability to capture and interpret complex error patterns. Notably, Llama3.1-8b also performed well, showcasing the effectiveness of our framework in helping smaller models understand data errors.

\begin{figure*}[t]
\vspace*{-0.1in}
\centering
\subfigure[Hospital]{\raisebox{-0.1cm}{\includegraphics[width=0.3\linewidth]{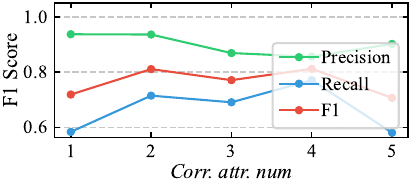}}}
\subfigure[Flights]{\raisebox{-0.1cm}{\includegraphics[width=0.3\linewidth]{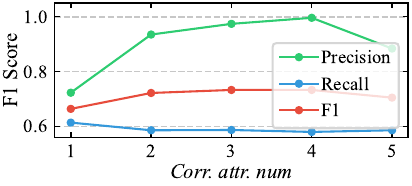}}}
\subfigure[Beers]{\raisebox{-0.1cm}{\includegraphics[width=0.3\linewidth]{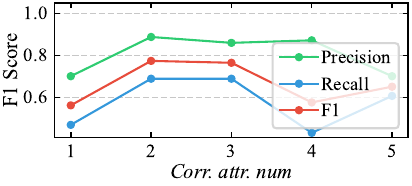}}}
\vspace*{-0.1in}
\\
\subfigure[Rayyan]{\raisebox{-0.1cm}{\includegraphics[width=0.3\linewidth]{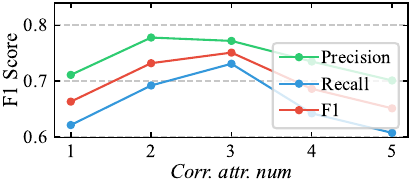}}}
\subfigure[Billionaire]{\raisebox{-0.1cm}{\includegraphics[width=0.3\linewidth]{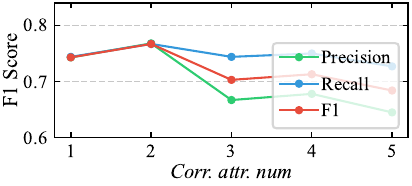}}}
\subfigure[Movies]{\raisebox{-0.1cm}{\includegraphics[width=0.3\linewidth]{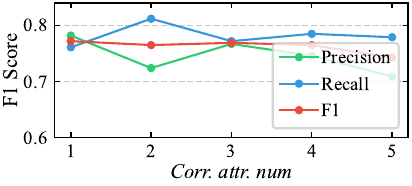}}}
\vspace*{-0.1in}
\caption{Error detection performance under different related attribute numbers.}
\label{fig:rel_attr_num}
\vspace{-0.18in}
\end{figure*}

\subsubsection{Effect of correlated attribute number}
Fig.~\ref{fig:rel_attr_num} illustrates how the number of correlated attributes affects ED performance across six datasets. Hospital and Flights maintain high precision with 2-4 attributes, Movies show stable performance across different attribute numbers, Billionaire performs best with 2 attributes, and Beers and Rayyan display more variability. Optimal performance is generally achieved with 2-3 related attributes, providing sufficient context while minimizing noise. Using only 1 attribute results in poor performance due to insufficient context, while including 5 attributes degrades performance due to increased noise and complexity. 
\subsubsection{Effect of clustering methods}
To further explore the effects of clustering methods, we adopt three different clustering methods, e.g., random sampling, Agglomerative Clustering~\cite{kaufman2009finding}, and $k$-means~\cite{lloyd1982kmeans} methods on Flights, Billionaire, and Movies.
Table~\ref{tab:cluster_methods} demonstrates that two clustering methods generally perform better than random sampling. Specifically, the $k$-means clustering method is more robust across various datasets, while Agglomerative Clustering can outperform in certain scenarios. Notably, for the Flights dataset, which contains simpler errors, there is less performance disparity between the methods. However, for the Billionaire and Movies datasets, which present more complex errors, the performance gap becomes more pronounced, highlighting the effectiveness of the clustering approaches over random sampling.

\begin{table}[t]
\small
\caption{Performance with different clustering methods.}
\label{tab:cluster_methods}
\renewcommand{\arraystretch}{1.15}
\setlength{\tabcolsep}{0.4mm} 
\centering
\begin{tabular}{|c|ccc|ccc|ccc|}
\hline
\multicolumn{1}{|c|}{\multirow{2}{*}{\makecell[c]{\textbf{Clustering} \\ \textbf{method}}}} & \multicolumn{3}{c|}{\textbf{Flights}} & \multicolumn{3}{c|}{\textbf{Billionaire}} & \multicolumn{3}{c|}{\textbf{Movies}} \\ \cline{2-10}
 & Prec & Rec & F1 & Prec & Rec & F1 & Prec & Rec & F1 \\
\hline
Random & 0.875 & \textbf{0.599} & 0.702 & 0.510 & 0.571 & 0.539 & 0.368 & 0.741 & 0.491 \\ \hline 
AGC & \textbf{0.946} & 0.580 & 0.719 & 0.668 & 0.762 & 0.718 & 0.589 & 0.720 & 0.632 \\ \hline 
$k$-Means & 0.935 & 0.586 & \textbf{0.722} & \textbf{0.768} & \textbf{0.767} & \textbf{0.767} & \textbf{0.724} & \textbf{0.812} & \textbf{0.765} \\ \hline
\end{tabular}
\vspace{-0.18in}
\end{table}

\subsection{Different Error Scenarios}
\begin{figure}[t]
\center
\vspace{-0.05in}
\includegraphics[width=0.94\linewidth]{time_legend.pdf} \\
  \includegraphics[width=0.96\linewidth]{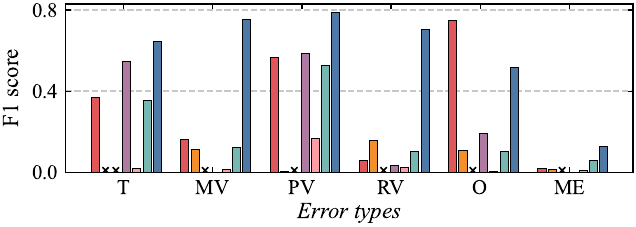}
  \vspace{-0.1in}
  \caption{Performance v.s. error types.}
 \label{fig:error_type}
 \vspace*{-0.18in}
\end{figure}

As shown in Fig.~\ref{fig:error_type}, we evaluate the F1 score on the Beers dataset across previous five error types selected from the original data, and corresponding error rates are shown previously. Especially,  mixed errors (ME) involving at least 3 former types with 0.49\% error rate, and `x' denotes zero value.
Although specialized methods like Nadeef (for rule violations) and dBoost (for outliers) excel in their specific scenarios, they require human-defined criteria. 
Except for RV error condition, ZeroED generally performs better.
ZeroED and FM\_ED show more robust performance than non-LLM baselines, particularly in mixed error scenarios, indicating the usage of LLM in better managing interrelated errors.


\section{Related Work}\label{sec:related_work}
This section provides an overview of existing ED methods, including those relying on external expertise, labeled errors, and LLMs.
Most prior ED methods depend on pre-defined manual criteria~\cite{Abiteboul1995Foundations,Beskales10sampling,Chu13holistic,Ebaid13nadeef,Fan08Cfds,Chu15Katara,Rekatsinas17holoclean,miao21influence}, typically developing heuristic strategies to find violations.
Manual label-based algorithms aim to provide more comprehensive detection by leveraging ML models trained on (weakly) labeled errors~\cite{Heidari19holodetect,Pham21spade,raha19mahdavi,Neutatz19ed2}.
However, these methods often require substantial human effort, and frequently rely on predefined statistical or simple lexical features, limiting their adaptability to diverse datasets and domains.

The emergence of LLMs such as ChatGPT~\cite{openai2024gpt4,openai24chatgpt}, Llama~\cite{touvron2023llama, dubey2024llama3herdmodels}, and Qwen~\cite{bai2023qwentechnicalreport,yang2024qwen2technicalreport}, has revolutionized current applications, demonstrating impressive capabilities in in-context learning~\cite{dong2023survey,min2022metaicl, liu2023moelora,wei24iterclean}, adherence to instructions~\cite{ouyang2022training,wei2022finetuned,xu2024large}, and chain-of-thought reasoning~\cite{wei2023chainofthought, zhao2024recommender, Ji23hallucinationsurvey}. 
Existing LLM-based tabular data error detection methods primarily focus on creating prompts~\cite{Narayan22Foundationwrangle} or fine-tuning LLMs for various table-related tasks~\cite{li2023tablegpt,openai24chatgpt,Huang2022TowardsRI,xu2024multi}.
They often examine tuples in isolation, lacking the necessary data contexts to detect complex errors.
In contrast, ZeroED uses LLMs for \emph{zero-shot} and holistic error labeling, meanwhile enhancing important feature representation, error labeling, and training data construction through LLM reasoning. This combination of both strengths, offers an effective and efficient ED framework.


\vspace{-0.05in}
\section{Conclusion}\label{sec:conclusions}
In this paper, we present ZeroED, a novel hybrid framework that combines LLM reasoning with previous ML pipelines via zero-shot prompting. ZeroED enhances feature representation through LLM-derived error-checking criteria features, meanwhile using efficient sampling and LLM-deduced guidelines to achieve holistic and zero-shot labeling with limited token cost. Training data quality is also improved through label propagation and error augmentation with verification. Experimental results demonstrate the superior performance and efficiency of ZeroED as a practical solution.

\textbf{Acknowledgements}.
This work is partly supported by the National Key R\&D Program (No. 2024YFB3908401), the National NSFC (No. 62372404), the Fundamental Research Funds for the Central Universities (No. 226-2024-00030), the Leading Goose R\&D Program of Zhejiang (No. 2024C01109), Research Impact Fund (No.R1015-23), Collaborative Research Fund (No.C1043-24GF), Hong Kong ITC Innovation and Technology Fund Midstream Research Programme for Universities Project (No.ITS/034/22MS), Huawei (Huawei Innovation Research Program, Huawei Fellowship), Tencent (CCF-Tencent Open Fund, Tencent Rhino-Bird Focused Research Program), Ant Group (CCF-Ant Research Fund), Alibaba (CCF-Alimama Tech Kangaroo Fund No. 2024002), and Kuaishou. Xiaoye Miao and Xiangyu Zhao are the corresponding authors of the work.



\bibliographystyle{IEEEtran}
\bibliography{6-References}

\end{document}